\documentclass[final]{cvpr}

\usepackage{times}
\usepackage{epsfig}
\usepackage{graphicx}
\usepackage{amsmath}
\usepackage{amssymb}

\usepackage{booktabs}
\usepackage{color}
\usepackage{multirow}
\usepackage{subcaption}
\usepackage{breqn}
\usepackage{textcomp}
\usepackage{multicol}

\usepackage{systeme}
\usepackage[ruled,vlined]{algorithm2e}

\definecolor{RED}{rgb}{0.5,0.1,0}
\definecolor{vert}{rgb}{0.1,0.5,0.2}
\definecolor{BLUE}{rgb}{0,0.2,0.6}
\definecolor{GREEN}{rgb}{0,0.6,0}
\definecolor{PURPLE}{rgb}{0.69,0,0.8}
\definecolor{BLACK}{rgb}{0.2,0.2,0.2}

\definecolor{Green}{rgb}{0.6,0.8,0.6}
\definecolor{Blue}{rgb}{0.1,0.3,0.6}
\definecolor{Orange}{rgb}{0.9,0.7,0.6}

\usepackage{amsmath,amsfonts,bm}

\def\eqref#1{equation~\ref{#1}}

\def\1{\bm{1}}

\def\vc{{\bm{c}}}

\def\vf{{\bm{f}}}

\def\vx{{\bm{x}}}

\def\mA{{\bm{A}}}
\def\mB{{\bm{B}}}

\def\mF{{\bm{F}}}

\def\mY{{\bm{Y}}}

\DeclareMathAlphabet{\mathsfit}{\encodingdefault}{\sfdefault}{m}{sl}
\SetMathAlphabet{\mathsfit}{bold}{\encodingdefault}{\sfdefault}{bx}{n}
\newcommand{\tens}[1]{\bm{\mathsfit{#1}}}

\def\tB{{\tens{B}}}

\def\sR{{\mathbb{R}}}

\usepackage[pagebackref=true,breaklinks=true,colorlinks,bookmarks=false]{hyperref}

\def\cvprPaperID{4339} 
\def\confYear{CVPR 2021}

\begin{document}

\title{Deep Miner: A Deep and Multi-branch Network which Mines Rich\\ and Diverse Features for Person Re-identification}

\author{Abdallah Benzine$^{1}$, Mohamed El Amine Seddik$^{2}$, Julien Desmarais$^{1}$\\
$^{1}$Digeiz AI Lab \\
$^{2}$Ecole Polytechnique\\
}


\maketitle



\begin{abstract}
Most recent person re-identification approaches are based on the use of deep convolutional neural networks (CNNs).  These networks, although effective in multiple tasks such as classification or object detection, tend to focus on the most discriminative part of an object rather than retrieving all its relevant features. This behavior penalizes the performance of a CNN for the re-identification task, since it should identify diverse and fine grained features. It is then essential to make the network learn a wide variety of finer characteristics in order to make the re-identification process of people effective and robust to finer changes. 

In this article, we introduce Deep Miner, a method that allows CNNs to ``mine'' richer and more diverse features about people for their re-identification. Deep Miner is specifically composed of three types of branches: a Global branch (G-branch), a Local branch (L-branch) and an Input-Erased branch (IE-branch). G-branch corresponds to the initial backbone which predicts global characteristics, while L-branch retrieves part level resolution features. The IE-branch for its part, receives partially suppressed feature maps as input thereby allowing the network to ``mine'' new features (those ignored by G-branch) as output. For this special purpose, a dedicated suppression procedure for identifying and removing features within a given CNN is introduced. This suppression procedure has the major benefit of being simple, while it produces a model that significantly outperforms state-of-the-art (SOTA) re-identification methods. Specifically, we conduct experiments on four standard person re-identification benchmarks and witness an absolute performance gain up to 6.5\% mAP compared to SOTA.

\end{abstract}

\section{Introduction}
\begin{figure}[t!]
  \centering
  \includegraphics[width=0.9\linewidth]{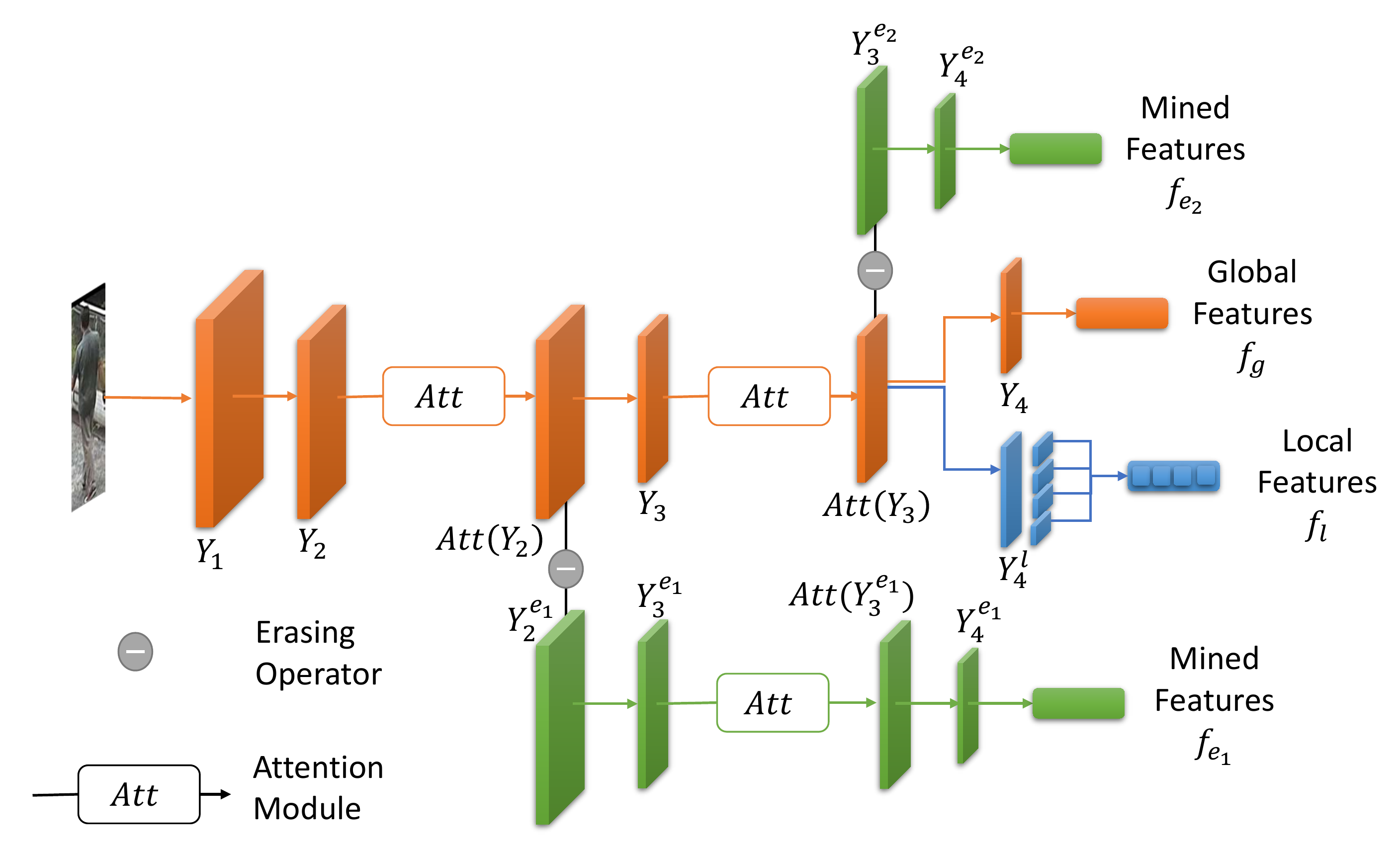}
\caption{\textbf{Deep Miner Model Architecture.} Given a standard CNN backbone, several branches are created to enrich and diversify the features for the purpose of person re-identification. The proposed Deep Miner model is made of three types of branches: \textbf{(i)} The main branch G (in {\color{Orange} \textbf{orange}}) is the original backbone and predicts the standard \textit{global features} $\vf_g$; \textbf{(ii)} Several Input-Erased (IE) branches (in {\color{Green} \textbf{green}}) that takes as input erased feature maps and predict \textit{mined features} $\vf_{e_1}$ and $\vf_{e_2}$; \textbf{(iii)} The local branch (in {\color{Blue} \textbf{blue}}) that outputs local features $\vf_l$ and in which a uniform partition strategy is employed for \textit{part level feature} resolution as proposed by \cite{xie2020learning}. In the global branch, and the bottom IE Branch, attention modules are used in order to  improve their feature representation.}
\label{fig:deep_mina}
\end{figure}

\begin{figure}[htbp]
    \centering

        \begin{subfigure}[t]{0.15\linewidth} 
        \centering \includegraphics[width=\textwidth]{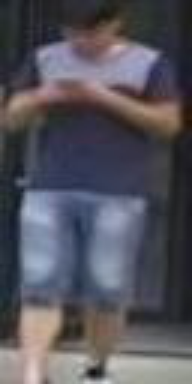}
    \end{subfigure}
   ~
    \begin{subfigure}[t]{0.15\linewidth} 
        \centering \includegraphics[width=\textwidth]{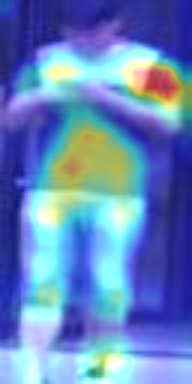}
    \end{subfigure}
   ~
        \begin{subfigure}[t]{0.15\linewidth} 
        \centering \includegraphics[width=\textwidth]{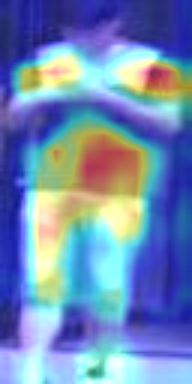}
    \end{subfigure}
   ~
          \begin{subfigure}[t]{0.15\linewidth} 
        \centering \includegraphics[width=\textwidth]{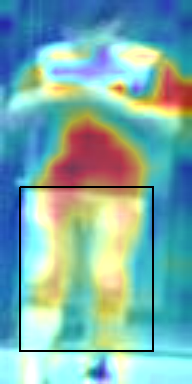}
    \end{subfigure}
   ~
             \begin{subfigure}[t]{0.15\linewidth} 
        \centering \includegraphics[width=\textwidth]{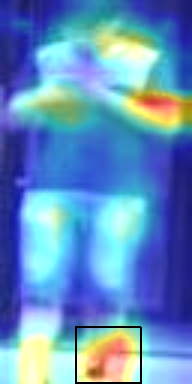}
    \end{subfigure}
    
    \begin{subfigure}[t]{0.15\linewidth} 
        \centering \includegraphics[width=\textwidth]{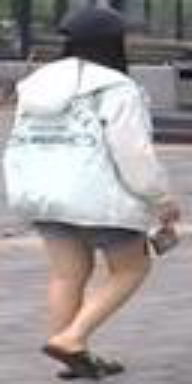}
    \end{subfigure}
    ~
    \begin{subfigure}[t]{0.15\linewidth} 
        \centering \includegraphics[width=\textwidth]{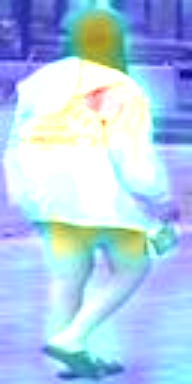}
    \end{subfigure}
~
        \begin{subfigure}[t]{0.15\linewidth} 
        \centering \includegraphics[width=\textwidth]{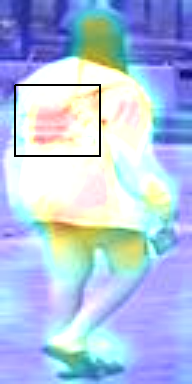}
    \end{subfigure}
  ~
          \begin{subfigure}[t]{0.15\linewidth} 
        \centering \includegraphics[width=\textwidth]{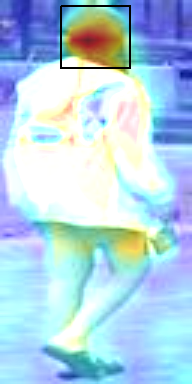}
    \end{subfigure}
    ~
             \begin{subfigure}[t]{0.15\linewidth} 
        \centering \includegraphics[width=\textwidth]{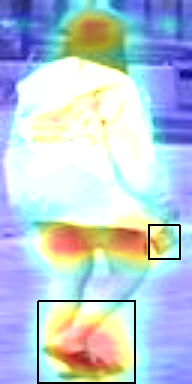}
    \end{subfigure}

    \begin{subfigure}[t]{0.15\linewidth} 
        \centering \includegraphics[width=\textwidth]{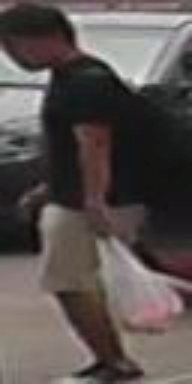}
        \caption*{Input}
    \end{subfigure}
   ~
    \begin{subfigure}[t]{0.15\linewidth} 
        \centering \includegraphics[width=\textwidth]{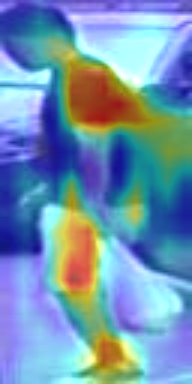}
        \caption*{Global}
    \end{subfigure}
    \texttildelow 
        \begin{subfigure}[t]{0.15\linewidth} 
        \centering \includegraphics[width=\textwidth]{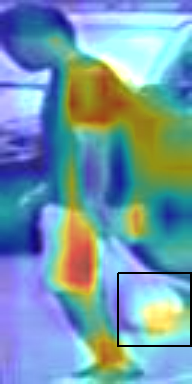}
        \caption*{IE 1}
    \end{subfigure}
    ~
          \begin{subfigure}[t]{0.15\linewidth} 
        \centering \includegraphics[width=\textwidth]{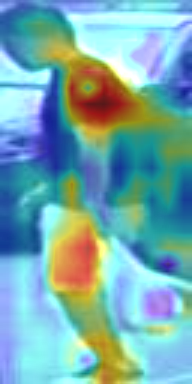}
        \caption*{IE 2}
    \end{subfigure}
  ~
             \begin{subfigure}[t]{0.15\linewidth} 
        \centering \includegraphics[width=\textwidth]{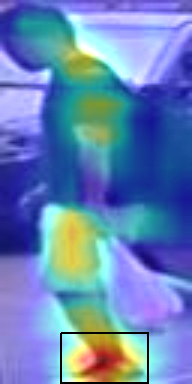}
        \caption*{Local}
    \end{subfigure}
\caption{\textbf{Feature visualization for three examples.} Warmer color denotes higher value. The global branch (second column) focuses only on some features but ignores other important ones. Thanks to the Input Erased branches (third and fourth columns), Deep Miner discovers new important features (localized by the black boxes). For instance, in the first row, the IE-branches are more attentive to the person pant.  In the second row, they discover some patterns on the coat and get attentive to its cap. In the third row, they find out the plastic bag. The local branch (last column) helps the network to focus on local features such as the shoes of the subject or the object handled by the subject in the second row.}
\label{fig:deep_miner_viss}

\end{figure}

In recent years, person re-identification (Re-ID) has attracted major interest due to its important role in various computer vision applications: video surveillance, human authentication, human-machine interaction etc. The main objective of person Re-ID is to determine whether a given person has already appeared over a network of cameras, which technically implies a robust modelling of the global appearance of individuals. The Re-ID task is particularly challenging because of significant appearance changes -- often caused by variations in the background, the lightening conditions, the body pose and the subject orientation w.r.t. the recording camera.

In order to overcome these issues, one of the main goals of person Re-ID models is to produce rich representations of any input image for person matching. Notably, CNNs are known to be robust to appearance changes and spatial location variations, as their global features are invariant to such transformations. Nevertheless, the aforementioned global features are prone to ignore detailed and potentially relevant information for identifying specific person representations. To enforce the learning of detailed features, attention mechanisms and aggregating global part-based representations were introduced in the literature, yielding very promising results \cite{chen2019mixed, chen2019abd, sun2018beyond}. Specifically, attention mechanisms allow to reduce the influence of background noise and to focus on relevant features, while part-based models divide feature maps into spatial horizontal parts thereby allowing the network to focus on fine-grained and local features.

Despite their observed effectiveness in various tasks, these approaches do not provide ways to enrich and diversify an individual's representation. In fact, deep learning models are shown to exhibit a biased learning behavior \cite{chen2019hybrid, chen2019energy, tamaazousti2019universal}; in the sense that they retrieve sufficiently partial attributes concepts which contribute to reduce the training loss over the seen classes, rather than learning all-sided details and concepts. Basically, deep networks tend to focus on surface statistical regularities rather than more general abstract concepts. This behavior is problematic in the context of re-identification, since the network is required to provide the richest and most diverse possible representations.\\

In this paper, we propose to address this problem by adding Input Erased Branches (IE-Branch) into a standard backbone. Precisely,  an IE-Branch takes partially removed feature maps as input in the aim of producing (that is ``mining'') more diversified features as output (as depicted in Figure \ref{fig:deep_miner_viss}). In particular, the removed regions correspond quite intuitively to areas where the network has strong activations and are determined by a simple suppression operation (see subsection \ref{ss:ie_branch}).  The proposed Deep Miner model is therefore made as the combination of IE-branches with local and global branches. The multi-branch architecture of Deep Miner is depicted on Figure \ref{fig:deep_mina}.\\

The main contributions brought by this work may be summarized in the following items: $\textbf{(i)}$ We provide a multi-branch model allowing the mining of rich and diverse features for people re-identification. The proposed model includes three types of branches: a Global branch (G-branch), a Local branch (L-branch) and an Input-Erased Branch (IE-Branch); the latter being responsible of mining extra features;
$\textbf{(ii)}$ IE-Branches are constructed by adding an erase operation on the global branch feature maps, thereby allowing the network to discover new relevant features;
$\textbf{(iii)}$ Extensive experiments were conducted on Market1501 \cite{zheng2015scalable}, DukeMTMC-ReID \cite{ristani2016performance}, CUHK03 \cite{li2014deepreid} and MSMT17\cite{wei2018person}. We demonstrate that our model significantly outperforms the existing SOTA methods on all benchmarks.

\section{Related Work}
There has been an extensive amount of works around the problem of people re-identification. This section particularly recalls the main advances and techniques to tackle this task, which we regroup subsequently in terms of different approaches:

\textbf{Part-level features} which essentially pushes the model to discover fine-grained information by extracting features at a part-level scale. Specifically, this approach consists in dividing the input image into multiple overlapping parts in order to learn part-level features \cite{yi2014deep, li2014deepreid}. In the same vein, other methods of body division were also proposed; Pose-Driven Deep Convolutional (PDC) leverages human pose information to transform a global body image into normalized part regions, Part-based Convolution Baseline (PCB) \cite{sun2018beyond} learns part level features by dividing the feature map equally -- Essentially, the network has a 6-branch structure by dividing the feature maps into six horizontal stripes and an independent loss is applied to each strip. Based on PCB, stronger part based methods were notably developed \cite{zheng2019pyramidal, quan2019auto, wang2018learning}. However, theses division strategies usually suffer from misalignment between corresponding parts -- because of large variations in poses, viewpoints and scales. In order to avoid this misalignment, \cite{xie2020learning} suggest to concatenate the part-level feature vectors into a single vector and then apply a single loss to the concatenated vector. This strategy is more effective than applying individual loss to each part-level feature vector. As shall be seen subsequently, we particularly employ this strategy to learn the local branch of our Deep Miner model.


\textbf{Metric learning} methods contribute also to enhance the representations power of Re-ID features as notably shown in \cite{chen2018group, hermans2017defense, sun2017svdnet, ristani2016performance}. For instance, the batch hard triplet loss introduced in \cite{hermans2017defense} retrieves the hardest positive and the hardest negative samples for each pedestrian image in a batch. Moreover, the soft-margin triplet loss\footnote{This loss function is used for our Deep Miner model training.} \cite{hermans2017defense} extends the hard triplet loss by using the \textsc{softplus} function to remove the margin hyper-parameter. Finally, authors in \cite{shen2018deep} improve the training and testing processes by gallery-to-gallery affinities and through the use of a group-shuffling random walk network. 

\textbf{Attention Modules} were also suggested to improve the feature representation of Re-ID models \cite{chen2019mixed, chen2019abd, hou2019interaction, tay2019aanet, si2018dual, li2018harmonious, xu2018attention}. Specifically, a dual attention matching network with inter-class and intra-class attention module that captures
the context information of video sequences was proposed in \cite{si2018dual}. In \cite{li2018harmonious}, a multi-task  model jointly learns soft pixel-level and hard region-level attention to improve the discriminative feature representations. In \cite{xu2018attention}, the final feature embedding is obtained by combining global and part features through the use of pose information to learn attention masks.\\

In the aim of learning fine grained features, a series of works focus on using erasing methods -- in various contexts beyond the Re-ID paradigm. We briefly recall the main works following this approach in the following paragraph and we particularly emphasize that our proposed Deep Miner model relies on a feature erasing approach.

\textbf{Feature erasing methods} were commonly used for weakly-supervised object localization \cite{zhang2018adversarial,choe2019attention, benassou2020hierarchical}. Technically, these methods rely on erasing the most discriminative part to encourage the CNN to detect other parts of an object. Our Deep Miner Model relies on the same principal whereas it fundamentally differs in terms of the targeted purpose and from the implementation standpoint. Indeed, Deep Miner aims to enrich the feature representation of deep neural networks in order to properly distinguish a person from another one. Similar to the proposed Deep Miner model, authors in \cite{chen2020salience} introduce a Salient Feature Extraction unit which suppresses the salient features learned in the previous cascaded stage thereby extracting other potential salient features. Still, our method differs from \cite{chen2020salience} through different aspects: $\textbf{(i)}$ the erasing operation used in Deep Miner is conceptually simpler -- since it consists in averaging and thresholding operations yielding an erasing mask -- while being very efficient in terms of outcome; $\textbf{(ii)}$ the obtained erasing mask is then multiplied by features maps of the initial CNN backbone allowing to create an Input-Erased branch that will discover new relevant features; $\textbf{(iii)}$ by simply incorporating such simple Input-Erased branches into the standard \texttt{Resnet50} backbone, Deep Miner achieves the same \textit{mAP} in Market1501 as \cite{chen2020salience} which include many other complex modules (e.g. attention modules); $\textbf{(iv)}$ Deep Miner contains a local branch with part level resolution that is absent in \cite{chen2020salience} and the two models differ in the used attention module; $\textbf{(v)}$ finally, \cite{chen2020salience} integrates a non-local multistage feature fusion which we found unnecessary for Deep Miner to achieve high re-identification performance.

\section{Proposed Method}
This section presents in more details our proposed method. As previously discussed, Deep Miner aims to enable the learning of more rich and fine grained features in the context of person re-identification. Specifically, Deep Miner relies on a given CNN backbone (e.g., \texttt{Resnet50}), and enriching it with new branches (each of them is described subsequently) to allow the network to mine richer and more diverse features.  The overall architecture of the network is described in Figure \ref{fig:deep_mina}.

\subsection{Global Branch}
The Global branch (G-branch) corresponds to a standard CNN backbone like \texttt{Resnet50}. This backbone is composed of $B$ convolutional blocks\footnote{As a standard residual network block.} $\tB_i$ for $i\in [B]$\footnote{The notation $[n] = \{1,\ldots,n\}$.}. The output of each block is denoted $\mY_i$. We particularly apply a global max pooling to the last convolutional layer (with stride being set to $1$) yielding to an output vector $\vf_g$. The global feature representation of a person is then obtained as a linear transformation of $\vf_g$.

\subsection{Input Erased Branch}
\label{ss:ie_branch}
\begin{figure}[t!]
  \centering
  \includegraphics[width=0.9\linewidth]{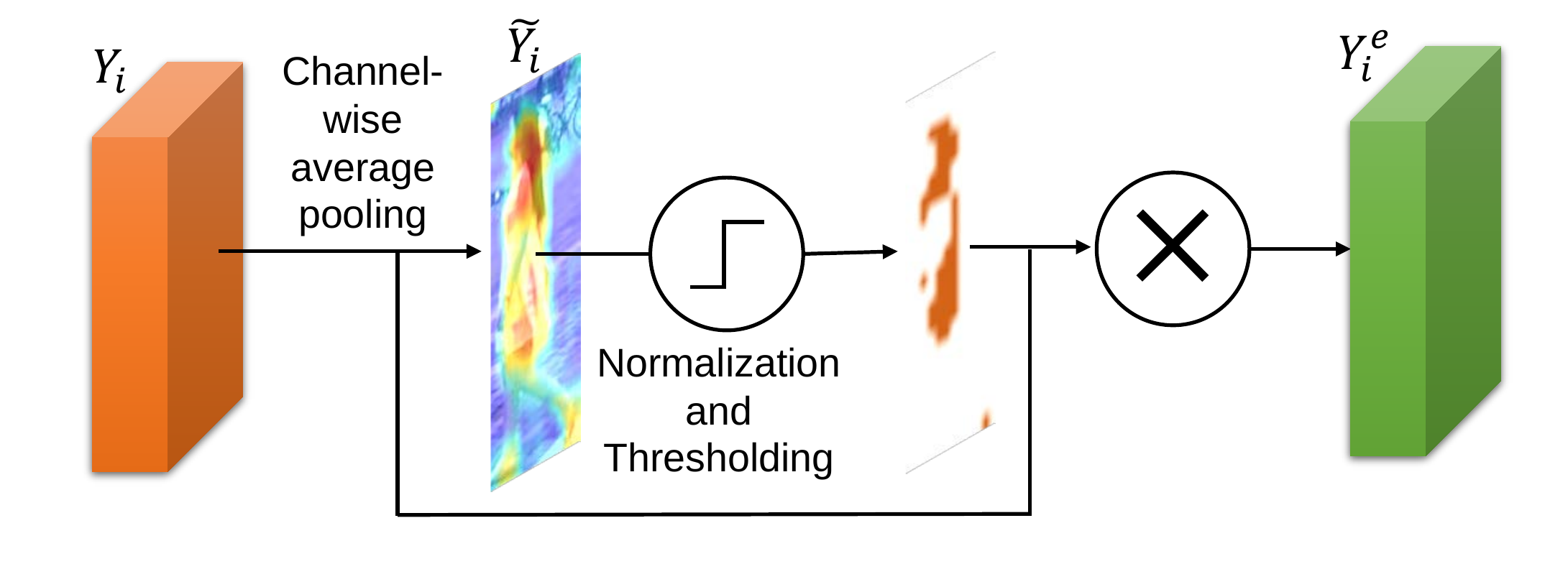}
\caption{Erasing Operation (\textit{ErO}).}
\label{fig:seo}
\end{figure}
As we discussed earlier, recent works \cite{chen2019hybrid, chen2019energy,tamaazousti2019universal} have  demonstrated that CNNs tend to focus only on the most discriminative parts of an image. In the context of person re-identification, this behavior is problematic since the network may not use important information to predict valuable features to increase person identification. 

To this end though, we propose to add new branches to the initial backbone network to mine a larger diversity of features from the images (yielding our Deep Miner proposed method). Specifically, we add Input-Erased branches (IE-branches) to the initial backbone. An IE-Branch can be added after any convolutional block $\tB_i$\footnote{If an attention module is used after block $\tB_i$, the IE-branch is added after this attention module.} of the global branch and provide a feature vector in the same way as the global branch. The convolutional block $\tB_{i+1}$ of the G-branch takes as input the unerased feature maps $\mY_i$\footnote{If an attention module is used after block $\tB_i$, $\mY_i$ stands for the output of this attention module.}, while the corresponding block in the IE-branch takes as input a partially erased version of it.  An \textit{Erasing Operation (ErO)} (illustrated in Figure \ref{fig:seo}) is particularly applied to the feature maps $\mY_i$ in order to obtain the erased features denoted $\mY^e_{i}$. The \textit{ErO} operation consists first in compressing $\mY_i$ through channel-wise average pooling so to get the average features maps $\tilde{Y_i}$. A min-max normalization is then applied to $\tilde{Y}_i$ to obtain $\tilde{Y}_i^n=\textsc{Min-Max-Norm}(\tilde{Y}_i)$. Given a thresholding parameter $\tau\in [0,1]$, an erasing mask $M_i$ is computed as follows:
\begin{equation}
 M_i(x,y) = \left\{
    \begin{array}{ll}
        0, & \mbox{if } \tilde{Y}_i^n(x,y)>\tau  \\
       1, & \mbox{otherwise}
    \end{array}
\right.
\end{equation}
where $I(x,y)$ stands for the pixel intensity at position $(x,y)$ of a 2D map $I$. As such, the erasing mask $M_i$ is simply obtained through averaging and thresholding operations. Furthermore, the erased features $\mY^e_{i}$ are then obtained by element-wise multiplication between $M_i$ and each channel of $\mY_i$, i.e., $\mY^e_{i}(c) = M_i \odot \mY_i(c) $ for $c$ indexing the channels.

As depicted in Figure \ref{fig:deep_mina}, creating an IE branch $k$ at block $\tB_i$ of the backbone will result in an additional branch composed of $B - i$ convolutional blocks which we denote $\tB_j^{e_k}$ for $j\in \{i+1, \ldots,B\}$, each of them being identical to the corresponding convolutional block of the main branch (G-branch).  In Figure \ref{fig:deep_mina},  an IE-Branch $1$ is created after the block $\tB_2$ (bottom IE-Branch). This IE-Branch is composed of the blocks $\tB^{e_1}_3$ and $\tB^{e_1}_4$ which have the same layers architecture as $\tB_3$ and $\tB_4$ and are initialised with same pretrained weights. If an attention module (subsection \ref{ss:attention_module}) is added to the backbone, the same attention module is added to the IE-branch. In the same way, another IE-Branch is created after block  $\tB_3$. We stress however that the weights are not shared between the different branches during training, so to let the network discover new features. 

Like in the global branch, we apply a global max pooling to the last convolutional layer of an IE Branch $k$ yielding to an output vector $\vf_{e_k}$ followed by a linear layer. 

Note that a similar erasing operation was already introduced in \cite{chen2020salience}. Nevertheless, the IE-branch in our proposed Deep Miner model differs from \cite{chen2020salience} in two fundamental ways: \textbf{(i)} the erasing mask is computed differently involving only a thresholding parameter $\tau$. Indeed, \textit{ErO} is simply based on average pooling and thresholding operations, while \cite{chen2020salience} uses a complex salient feature extractor that divides the features maps into several stripes,  the selector guiding each stripe to mine important information; \textbf{(ii)} the erasing operation is performed in a cascaded way at the end of the \texttt{Resnet50} backbone in \cite{chen2020salience} , while it is performed at different stages of the backbone in our Deep Miner model. This notably allows our model to mine new relevant features at different resolutions and semantic levels. 

\subsection{Local Branch}
\begin{figure}[t!]
  \centering
  \includegraphics[width=0.8\linewidth]{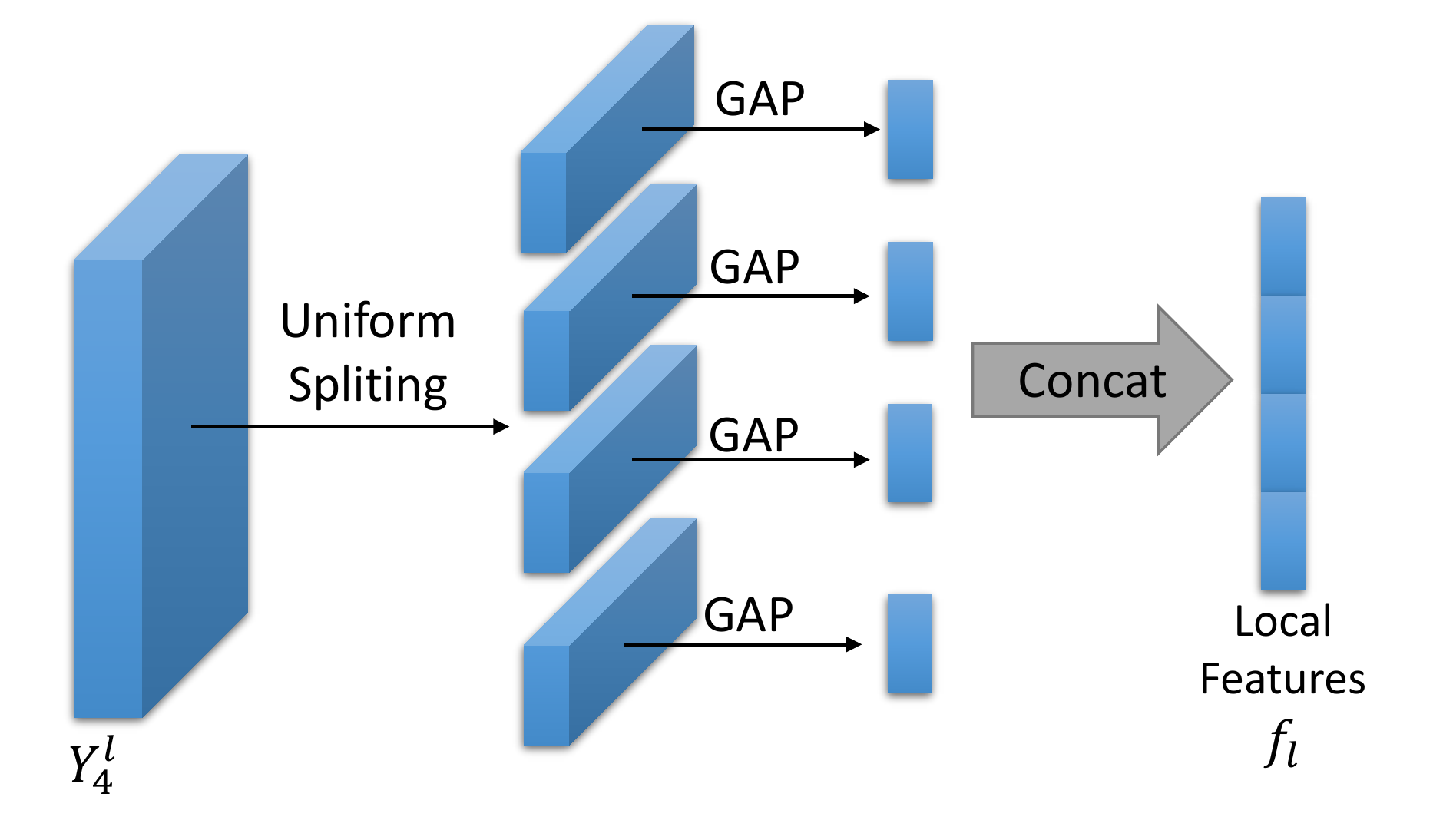}
\caption{Local Branch of Deep Miner.}
\label{fig:local_branch}
\end{figure}

While the IE-erased branches help the network to mine more diverse features, these branches are still based on the global appearance of a person. To help the network to mine local and more precise features, a local branch is added to Deep Miner. This branch is placed after the convolutional block $\tB_3$ and is composed of the convolutional block $\tB^l_4$ which have the same layers architecture as $\tB_4$ and is initialised with the same pretrained weights. Like in the IE-Branches, the weights are not shared between the different branches during training, so to let the network discover new localized features.

As illustrated in Figure \ref{fig:local_branch}, the local branch outputs features maps that are then partitioned into $4$ horizontal stripes. A global average pooling is then applied to each strip to obtain $4$ local features vectors. The $4$ local vectors are then concatenated yielding to an output vector $\vf_{l}$ as performed in \cite{xie2020learning}.

\subsection{Attention Modules}
\label{ss:attention_module}

Attention modules are commonly used in various deep learning application tasks and specifically in the context of person re-identification. The proposed Deep Miner model is also compatible with attention modules, which notably yield an enhancement of the model ability to retrieve more relevant features. To stress out the effectiveness of attention modules on the proposed method, we implement a simple attention module composed of a Spatial Attention Module (\textsc{Sam}) and a CHannel Attention Module (\textsc{Cham}) which we describe subsequently. Features maps $\mY_i$ are first processed by \textsc{Sam}, the result of which is then processed by \textsc{Cham}. The obtained features after \textsc{Sam} and \textsc{Cham} are denoted $\textsc{Att}(\mY_i)=\textsc{Cham}(\textsc{Sam}(\mY_i))$.

\begin{figure}[h!]
  \centering
  \includegraphics[width=0.9\linewidth]{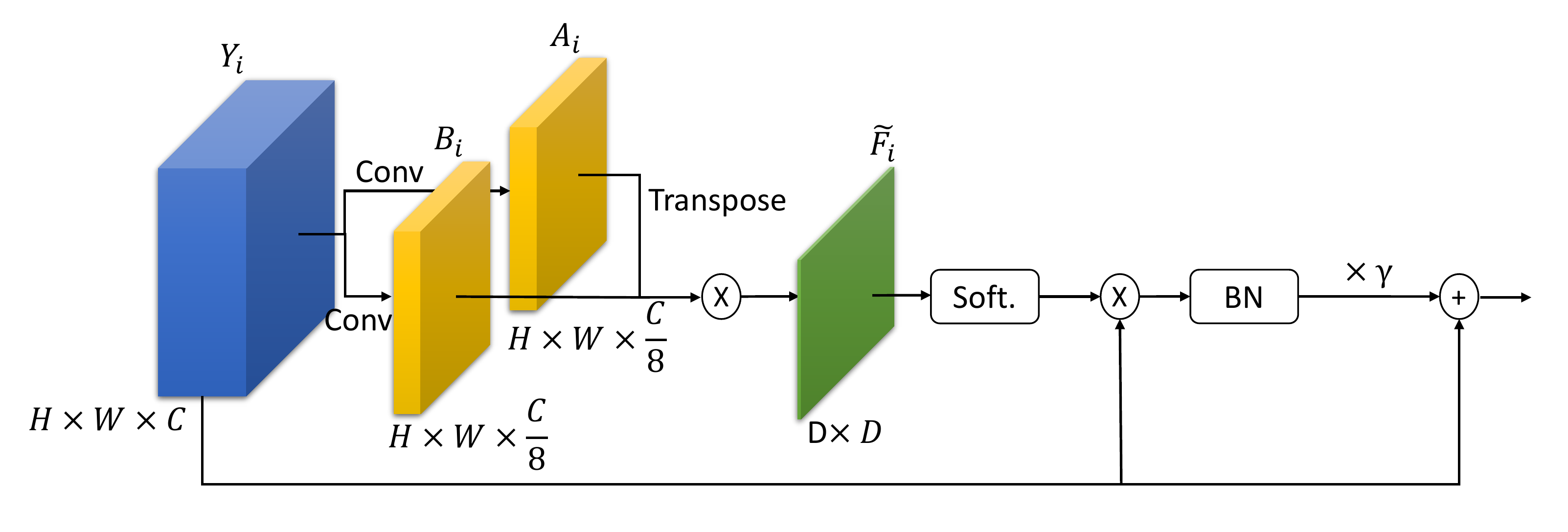}
\caption{Spatial Attention Module (\textsc{Sam}). }
\label{fig:sam}
\end{figure}

\textsc{Sam} focuses on the most relevant features within the spatial dimension which is essentially based on \cite{chen2019abd, xie2020learning}. Indeed, \textsc{Sam} captures and aggregates related features in the spatial domain. An illustration of this module is depicted in Figure \ref{fig:sam}. The input feature maps $\mY_i$ of dimension $H \times W \times C$ corresponding to the output of the convolutional block $\tB_i$ are fed into two convolutional layers to get two features maps $\mA_i$ and $\mB_i$ both of dimension $H \times W \times \frac{C}{8}$. The tensor $\mA_i$ is transposed and reshaped to shape $D \times C$ and $\mB_i$ is reshaped to $C \times D$, with $D=H\times W$. An affinity matrix $\tilde\mF_i = \mA_i^r \cdot \mB_i^r \in \sR^{D\times D}$ is computed by a matrix multiplication between the reshaped tensors $\mA_i$ and $\mB_i$ denoted respectively as $\mA_i^r$ and $\mB_i^r$. A Softmax activation is then applied to $\tilde\mF_i$ leading to $\mF_i$. After a reshaping operation, $\mY_i$ is multiplied (map-wise) by $\mF_i$ and a Batch Normalization layer is applied to the resulting tensor followed by a multiplication with a learnable scalar $\gamma$. This parameter adjusts the importance of the \textsc{Sam} transformation. The result is then added to the input $\mY_i$ to get $\textsc{Sam}(\mY_i)$, the resulting tensor of the Spatial Attention Module. 

\begin{figure}[h!]
  \centering
  \includegraphics[width=0.9\linewidth]{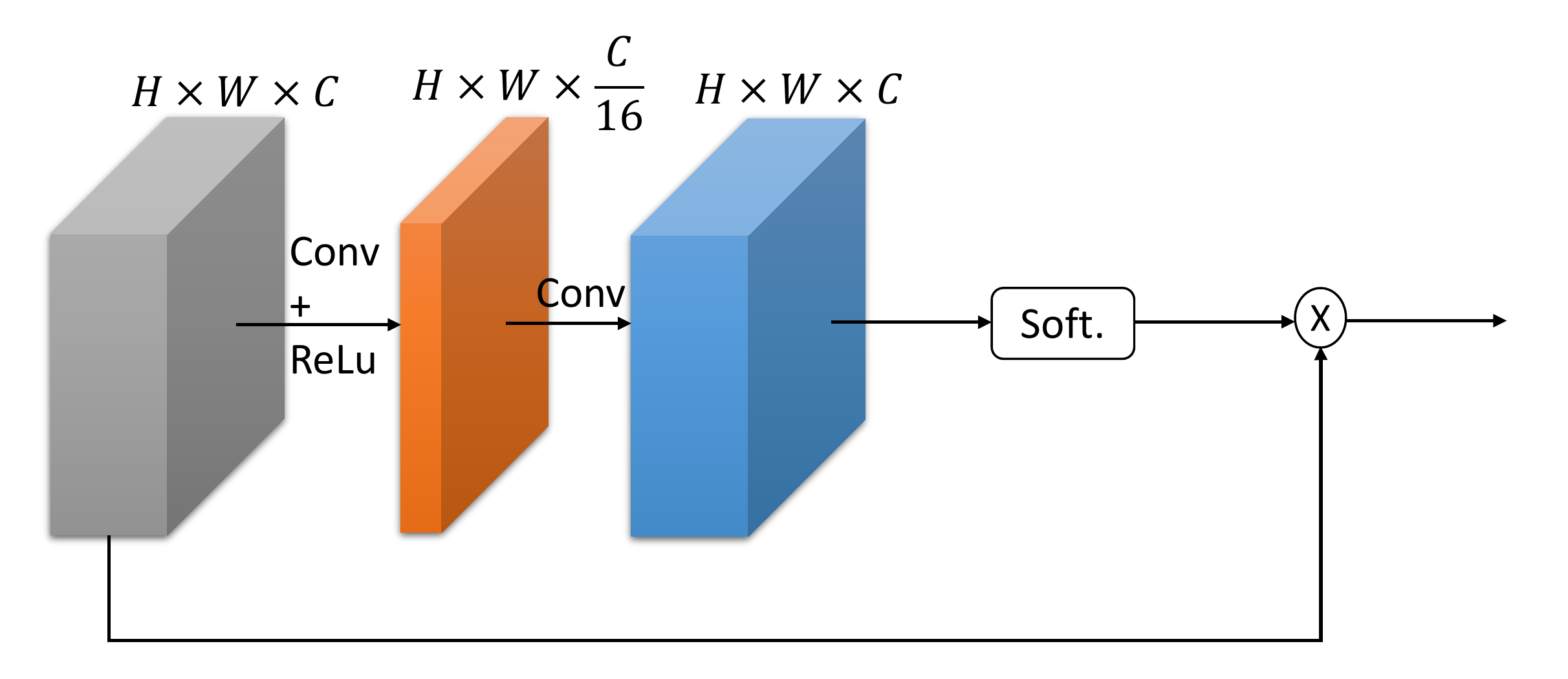}
\caption{CHannel Attention Module (\textsc{Cham}).}
\label{fig:cham}
\end{figure}

In contrast, \textsc{Cham} as illustrated in Figure \ref{fig:cham}, explores the correlation and the inter-dependencies between
channel features. It is specifically based on the Squeeze-and-Exitation block \cite{hu2018squeeze}. Besides, compared to \cite{hu2018squeeze}, the global average pooling at the beginning of the block is removed to preserve spatial information into the attention block. A convolutional layer is applied to $\textsc{Sam}(\mY_i)$ to obtain feature maps of size $H \times W \times \frac{C}{16}$ for which a second convolution layer is applied to obtain feature maps of size $H \times W \times C$. A Softmax activation is applied to the result that is then multiplied element-wise by $\textsc{Sam}(\mY_i)$ to obtain the result of \textsc{Cham}.

As depicted in Figure \ref{fig:deep_mina}, the attention module is placed right after the blocks $\tB_2$ and $\tB_3$ of the G-branch as well as after the block $\tB^{e_1}_3$ of IE-branch $1$ (bottom IE branch).


\subsection{Loss Functions}

Each branch $\texttt{Br}\in \{g,l,e_1, e_2, \ldots\}$ (among global branch, local branch and IE branches) outputs a corresponding feature vector $\vf_{\texttt{Br}}$. We apply the sames losses for each feature vector, i.e., an ID loss with label smoothing $\mathcal{L}_{\text{ID}}^{\texttt{Br}}$, a soft margin triplet loss $\mathcal L_{\text{triplet}}^{\texttt{Br}}$ and a center loss  $\mathcal L_{\text{center}}^{\texttt{Br}}$ yielding a global loss for each branch \texttt{Br} as 
\begin{equation}
    \mathcal L_{\texttt{Br}} = \mathcal{L}_{\text{ID}}^{\texttt{Br}} +  \mathcal L_{\text{triplet}}^{\texttt{Br}} + \lambda \mathcal L_{\text{center}}^{\texttt{Br}}
\end{equation}
where $\lambda > 0$ is an hyper-parameter ($\lambda=5\cdot 10^{-4}$ in all our experiments). The ID loss is specifically defined as
\begin{equation}
   \mathcal{L}_{\text{ID}}^{\texttt{Br}} = - \frac1N \sum_{i=1}^{N} q_i \log\left(p_i^{\texttt{Br}}\right)
\end{equation}
with $N$ standing for the number of samples, $p_i^{\texttt{Br}}$ denotes the predicted probability for identity $i$, while $q_i$ stands for the (ground-truth) smoothed label, and is defined as 
\begin{equation}
q_i = \left\{
    \begin{array}{ll}
        1- \epsilon\frac{N-1}{N}, & \mbox{if } i=y  \\
       \frac{\epsilon}{N}, & \mbox{otherwise}
    \end{array}
\right.
\end{equation}
where $y$ is the (hard) ground-truth identity and $\epsilon$ is a precision parameter ($\epsilon=0.1$ in practice) used to enforce the model to be less confident on the training set. Note that we apply the BNNeck \cite{luo2019bag} to the feature vector $\vf_{\texttt{Br}}$ right before the linear layer predicting the IDs probabilities.  

The soft margin triplet loss $\mathcal L_{\text{triplet}}^{\texttt{Br}}$ is employed to enhance the final ranking performance of the Re-ID model. $\mathcal L_{\text{triplet}}^{\texttt{Br}}$ is formally defined as 

\begin{dmath}
     \sum_{i=1}^{P} \sum_{k=1}^{K} \textsc{Softplus} ( \max_{\ell\in [K]} \left\Vert \vf_{\texttt{Br}}(\vx_k^i) - \vf_{\texttt{Br}}(\vx^i_\ell) \right\Vert_2 - \min_{ \substack{j\in [P]\setminus\{i\} \\ \ell\in [K] }} \left\Vert \vf_{\texttt{Br}}(\vx^i_k) - \vf_{\texttt{Br}}(\vx^j_\ell) \right\Vert_2 )
\end{dmath}
where $P$ stands for the number of identities per batch, $K$ is the number of samples per identity, $\vx^i_k$ is the $k$-th image of person $i$ and $\vf_{\texttt{Br}}(\vx^i_k)$ is the corresponding predicted feature vector (extracted among the branches of Deep Miner). 

Following \cite{luo2019bag}, the Center Loss $\mathcal L_{\text{center}}^{\texttt{Br}}$ is also considered for training our Deep Miner model. It simultaneously learns a center
for deep features of each identity and penalizes the distances
between the deep features and their corresponding identity
centers. As such, intra-class compactness is increased. The Center Loss is particularly defined as
\begin{equation}
    \mathcal L_{\text{center}}^{\texttt{Br}} = \frac{1}{2}\sum_{i=1}^P \sum_{k=1}^K \left\Vert \vf_{\texttt{Br}}(\vx_k^i)-\vc^i_{\texttt{Br}}\right\Vert^2_2 
\end{equation}
where $\vc^i_{\texttt{Br}}$ is the center of identity $i$ for the branch $\texttt{Br}$.

\section{Experiments}
\subsection{Experimental details} Deep Miner is implemented in PyTorch and trained on a single Nvidia GV100 GPU. All images are resized into $384\times 128$ pixels. Random Horizontal flipping and erasing are used during training. Each mini batch consists of $N = P \times n$ images where $P$ is the number of randomly selected identities and $n$ is the number of samples per identity. We take $P=16$ and $n=4$. We employ Adam as the optimizer with a warm-up strategy for the learning rate.  We spent $10$ epochs linearly increasing the learning rate from $3.5 \times 10^{-5}$ to $3.5 \times 10^{-4}$. Then, the learning rate is decayed to $3.5 \times 10^{-5}$
and $3.5 \times 10^{-6}$ at $40$-th epoch and $70$-th epoch respectively. The model is trained until convergence. The feature vector of an input image produced by Deep Miner corresponds the concatenation of the feature vectors predicted by each branch after application of the BNNeck. A \texttt{Resnet50} backbone is used in all experiments. 

\subsection{Evaluation Metrics}

To compare the performance of the proposed method with previous SOTA methods, the Cumulative Matching Characteristics (CMC) at \textit{Rank-1} and mean Average Precision (\textit{mAP}) are adopted. 
\subsection{Datasets}
We considered four standard re-identification benchmarks, the details of which are provided subsequently.

\textbf{Market1501} \cite{zheng2015scalable} consists of $1501$ identities and $32668$ images. $12936$ images of $751$ subjects are used for training while $19732$ images of $750$ subjects are used for testing with $3368$ query images and $16364$ gallery images. The images are shot by six cameras. The Deformable Part Model is used to generate the bounding boxes \cite{felzenszwalb2008discriminatively}.

\textbf{DukeMTMC-ReID} \cite{ristani2016performance} contains $36411$ images of $1404$ identities captured by more than $2$ cameras. The training subset contains $702$ identities with $16522$ images and the testing subset has other $702$ identities. The gallery set contains $17661$ images and the query set contains $2228$ images. 

\textbf{CUHK03} \cite{li2014deepreid} contains $1467$ identities and a total of  $14096$ labeled images and $14097$ detected captured by two camera views. $767$ identities are used for training and $700$ identities are used for testing. The labeled dataset contains $7368$ training images, $5328$ gallery and $1400$ query images for testing, while the detected dataset contains $7365$ images for training, $5332$ gallery, and $1400$ query images for test.

\textbf{MSMT17} \cite{wei2018person} is the largest and more challenging person Re-ID dataset. $4101$ identities and  $126441$ images are captured by a $15$-camera network ($12$ outdoor and $3$ indoor). Faster RCNN \cite{ren2016faster} is used to annotate the bounding boxes. 

\subsection{Ablation Studies}

To demonstrate the effectiveness of the IE and Local branches on the performance of Deep Miner, we incrementally evaluate each module on \textbf{Market-1501}. First, we evaluate the effect of the IE-branch. 

\begin{figure}[h!]
    \centering
    \includegraphics[width=0.34\textwidth]{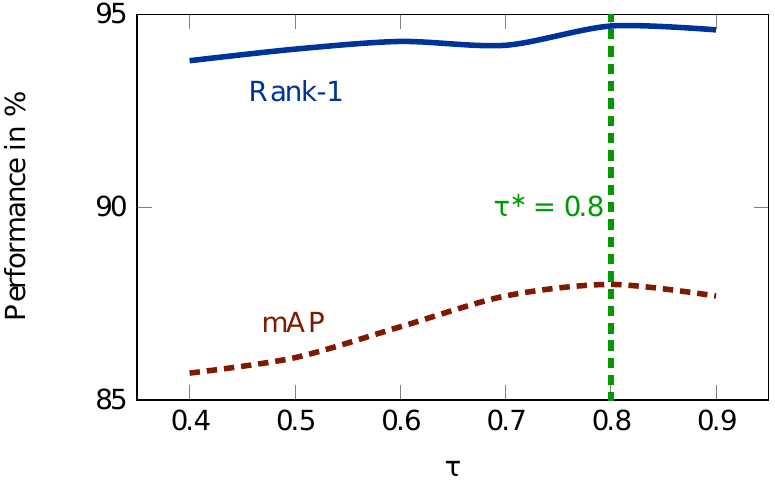}
    \caption{Influence of the erasing threshold $\tau$ with a single IE-Branch. The IE-Branch is placed after $\mB_3$,  the third \texttt{Resnet50} convolution block. The baseline (only global branch) has $94.2\%$ \textit{Rank-1} and $84.7\%$ \textit{mAP}.}
    \label{tab:erasing_th}
\end{figure}

\textbf{Erasing threshold:} We evaluate a model with a single IE branch and vary the erasing threshold (Figure \ref{tab:erasing_th}).  Compared to the baseline,  adding the IE Branch improves the \textit{mAP} with all the evaluated thresholds. Nevertheless, a careful choice of the optimal value of the threshold is needed for an optimal performance. With a low threshold, too much features are erased and the IE-Branch cannot mine new significant features. With a high threshold, no enough features are removed are the IE-branch does not discover new ones. The highest scores are obtained with $\tau=0.8$ with $94.7\%$ \textit{Rank-1} score and $88.0\%$ \textit{mAP} score.

\textbf{IE-Branch Position:} We now evaluate the importance of the position of the IE-branch (see Table \ref{tab:erasing_pos}). The IE-branch has a significant impact on the model performance, whatever its position in the CNN backbone. Still, the optimal scores are obtained at a particular position of the backbone. In fact, the best scores are obtained when the IE-branch is placed after the third convolution block with a $94.7\%$ \textit{Rank-1} and $88.0\%$ \textit{mAP}. 

\begin{table}[h!]
\centering
\footnotesize
\begin{tabular}{@{}lll@{}}
\toprule
IE Branch Position             & \textit{mAP} & \textit{Rank-1}     \\ \midrule
baseline (only global branch)    & $84.7$ & $94.2$    \\
After block 1                    & $87.0$  & $94.3$     \\
After block 2                    & $87.4$ & $94.6$    \\
After block 3                  & $\textbf{88.0}$ & $\textbf{94.7 }$    \\ \bottomrule
\end{tabular}
\caption{Influence of IE-Branch position ($\tau=0.8$).}
\label{tab:erasing_pos}
\end{table}

\textbf{Multiple IE-Branches:} In this paragraph we evaluate whether the incorporation of multiple IE-branches into Deep Miner improves the results compared to a single IE-branch. Table \ref{tab:multiple_ie_branches} shows that adding IE-branches after the convolution block $3$ and the convolutions block $1$ or $2$ improves the re-identification performance. Indeed, multiple IE-branches ensure that Deep Miner discovers new features at different semantics levels of the initial backbone. In particular, the optimal scores are obtained by adding IE-branches after the blocks $2$ and $3$. However, adding IE-branches after three different convolutional blocks does not yield a significant improvement compared to the $2$ IE-branches architecture. 

\begin{table}[h!]
\centering
\footnotesize
\begin{tabular}{@{}lll@{}}
\toprule
IE Branch Positions   & \textit{mAP} & \textit{Rank-1}  \\ \midrule
After block 3           & $88.0$  & $94.7$\\
After blocks 2 and 3   & $\textbf{88.5}$  & $\textbf{95.3}$\\
After blocks 1 and 3   & $88.3$ & $95.1$  \\
After blocks 1,2,3      & $88.2$ & $95.0$ \\ \bottomrule
\end{tabular}
\caption{Influence of multiple IE-Branches ($\tau=0.8$).}
\label{tab:multiple_ie_branches}
\end{table}

\textbf{Local Branch:} Now we evaluate the effect of adding a local branch. Table \ref{tab:local_branch} shows that adding this branch to the optimal Multiple IE-branches (obtained by the previous ablation study, i.e., adding IE-branches after blocks $2$ and $3$) significantly improves the \textit{mAP} ($89.4\%$ versus $88.5\%$).

\begin{table}[h!]
\centering
\footnotesize
\begin{tabular}{@{}lll@{}}
\toprule
Local Branch  & \textit{mAP} & \textit{Rank-1} \\ \midrule
No              & $88.5$ & $\textbf{95.3}$ \\
Yes            & $\textbf{89.4}$ & $95.2$\\ \bottomrule


\end{tabular}
\caption{Influence of the Local Branch.  IE-branches are placed after blocks $2$ and $3$ ($\tau=0.8$).}
\label{tab:local_branch}
\end{table}

\textbf{Attention Modules:} As previously discussed, attention modules are shown to increase the performance of modern person re-identification models. We therefore evaluate in this part the effect of such modules on Deep Miner. Table \ref{tab:attention_module} shows that the attention module described in Subsection \ref{ss:attention_module} helps the proposed model to mine relevant features with a $1.0\%$ absolute \textit{mAP} improvement. 

\begin{table}[h!]
\centering
\footnotesize
\begin{tabular}{@{}lll@{}}
\toprule
Attention Module  & \textit{mAP}  & \textit{Rank-1} \\ \midrule
No                 & $89.4$ & $95.2$ \\
\textsc{Sam}                 & $90.2$ & $95.5$ \\
\textsc{Cham}                 & $89.4$ & $95.5$ \\
\textsc{Sam} \& \textsc{Cham}            & $\textbf{90.4}$  & $\textbf{95.7}$ \\ \bottomrule
\end{tabular}
\caption{Influence of the Attention Module. IE-branches are placed after blocks $2$ and $3$ ($\tau=0.8$) and a Local branch is used.}
\label{tab:attention_module}
\end{table}

In the next section, we consider the optimal Deep Miner model obtained through this ablation studies which we compare to State-of-the-art methods in terms of \textit{mAP} and \textit{Rank-1} score. Specifically, we define the Deep Miner model as a \texttt{Resnet50} with two IE-branches placed after blocks $2$ and $3$, a local branch and attention modules (see Figure \ref{fig:deep_mina}).


\subsection{Comparison with State-of-the-art Methods}

\begin{table}[]
\centering
\footnotesize
\begin{tabular}{@{}l|l|ll@{}}
\toprule
Method                                                                 & Backbone    & \textit{mAP}            & \textit{Rank-1} \\ \midrule
Deep Miner & \texttt{Resnet50}    & \textbf{90.40} & \textbf{95.70 } \\ \midrule
$* +$ PLR OSNet \cite{xie2020learning}                                                              & \texttt{OSNet}       & 88.90          & 95.60  \\

\texttildelow  HOReID \cite{wang2020high} & \texttt{Resnet50} & 84.9 & 94.2 \\

$* +$ SCSN \cite{chen2020salience}                                                                   & \texttt{Resnet50}    & 88.50          & \textbf{95.70}  \\
 $*$ ABDNet \cite{chen2019abd}                                                                 & \texttt{Resnet50}    & 88.28          & 95.60  \\
$+$ Pyramid      \cite{zheng2019pyramidal}                                                          & \texttt{Resnet101}   & 88.20          & \textbf{95.70}  \\
DCDS \cite{alemu2019deep}                                                                   & \texttt{Resnet101}   & 85.80          & 94.81  \\
$* +$ MHN \cite{chen2019mixed}                                                                    & \texttt{Resnet50}    & 85.00          & 95.10  \\
$+$ MGN \cite{lin2019improving}                                                          & \texttt{Resnet50}    & 86.9           & 95.70  \\
BFE  \cite{dai2019batch}                                                                           & \texttt{Resnet50}    & 86.20          & 95.30  \\
$* +$CASN \cite{zheng2019re}                                                                   & \texttt{Resnet50}   & 82.80          & 94.40  \\
$* +$AANet \cite{tay2019aanet}                                                                 & \texttt{Resnet152}   & 83.41          & 93.93  \\
$*$ IANet \cite{hou2019interaction}                                                                 & \texttt{Resnet50}    & 83.10          & 94.40  \\
$* +$ VPM  \cite{sun2019perceive}                                                                  & \texttt{Resnet50}    & 80.80          & 93.00  \\
\texttildelow  PSE+ECN \cite{saquib2018pose}                                                               & \texttt{Resnet50}   & 80.50          & 90.40  \\
$+$ PCB+RPP \cite{sun2018beyond}                                                               & \texttt{Resnet50}    & 81.60          & 93.80  \\
$+$ PCB \cite{sun2018beyond}                                                                    & \texttt{Resnet50}    & 77.40          & 92.30  \\
\texttildelow  Pose-transfer \cite{liu2018pose}                                                         & \texttt{DenseNet169} & 56.90          & 78.50  \\
\texttildelow  SPReID  \cite{kalayeh2018human}                                                               & \texttt{Resnet152}  & 83.36          & 93.68  \\ \midrule

* RGA-SC \cite{zhang2020relation} & \texttt{Resnet50} & 88.4 & 96.1 \\
SNR \cite{jin2020style} & \texttt{Resnet50} & 84.70 & 94.40 \\
OSNet \cite{zhou2019omni} &\texttt{OSNet} & 84.9 & 94.8 \\
Tricks \cite{luo2019bag}                                                                 & \texttt{SEResNet101} & 87.30          & 94.60  \\
$*$ Mancs \cite{wang2018mancs}                                                                 & \texttt{Resnet50}    & 82.30          & 93.10  \\
PAN  \cite{zheng2018pedestrian}                                                                  & \texttt{Resnet50}    & 63.40          & 82.80  \\
SVDNet \cite{sun2017svdnet}                                                                 & \texttt{Resnet50}    & 62.10          & 82.30  \\ \bottomrule

\end{tabular}
\\ *Attention related, +Stripes Related, \texttildelow Pose or human pose related
\caption{Comparison with state-of-the-art person Re-ID methods
on the Market1501 dataset.}
\label{tab:market}
\end{table}

\begin{table}[]
\centering
\footnotesize
\begin{tabular}{ll|l|l}
\hline
Method                                                                 & Backbone    & \textit{mAP}   & \textit{Rank-1} \\ \hline
Deep Miner & \texttt{Resnet50}   & \textbf{81.80} & 91.20  \\ \hline
$* +$ PLR OSNet \cite{xie2020learning}                                 & \texttt{OSNet}       & 81.20 &\textbf{ 91.60}  \\
\texttildelow  HOReID \cite{wang2020high} & \texttt{Resnet50} & 75.60 & 86.90 \\
$* +$ SCSN \cite{chen2020salience}                                     & \texttt{Resnet50}    & 79.00 & 91.00  \\
$*$ ABDNet \cite{chen2019abd}                                          & \texttt{Resnet50}    & 78.60 & 89.00  \\
$+$ Pyramid      \cite{zheng2019pyramidal}                             & \texttt{Resnet101}   & 79.00 & 89.00  \\
DCDS \cite{alemu2019deep}                                              & \texttt{Resnet101}   & 75.50 & 87.50  \\
$* +$ MHN \cite{chen2019mixed}                                         & \texttt{Resnet50}    & 77.20 & 89.10  \\
$+$ MGN \cite{lin2019improving}                                        & \texttt{Resnet50}    & 78.40 & 88.70  \\
BFE  \cite{dai2019batch}                                               & \texttt{Resnet50}    & 75.90 & 88.90  \\
$* +$CASN \cite{zheng2019re}                                           & \texttt{Resnet50}    & 73.70 & 87.70  \\
$* +$AANet \cite{tay2019aanet}                                         & \texttt{Resnet50}   & 74.29 & 87.65  \\
$*$ IANet \cite{hou2019interaction}                                    & \texttt{Resnet50}    & 73.40 & 83.10  \\
$* +$ VPM  \cite{sun2019perceive}                                      & \texttt{Resnet50}    & 72.60 & 83.60  \\
\texttildelow  PSE+ECN \cite{saquib2018pose}                                      & \texttt{Resnet50}    & 75.70 & 84.50  \\
$+$ PCB+RPP \cite{sun2018beyond}                                       & \texttt{Resnet50}    & 69.20 & 83.30  \\
\texttildelow  Pose-transfer \cite{liu2018pose}                                   & \texttt{Densenet169} & 56.90 & 78.50  \\
\texttildelow  SPReID  \cite{kalayeh2018human}                                    & \texttt{Resnet152}   & 73.34 & 85.95  \\ \hline
SNR \cite{jin2020style} & \texttt{Resnet50} & 72.9 & 84.4 \\
OSNet \cite{zhou2019omni}                                              & \texttt{OSNet}       & 73.50  & 88.60   \\
Tricks \cite{luo2019bag}                                               & \texttt{SeResnet101} & 78.00 & 87.50  \\
$*$ Mancs \cite{wang2018mancs}                                         & \texttt{Resnet50}    & 71.80 & 84.90  \\
PAN  \cite{zheng2018pedestrian}                                        & \texttt{Resnet50}    & 51.51 & 71.59  \\
SVDNet \cite{sun2017svdnet}                                            & \texttt{Resnet50}    & 56.80 & 76.70  \\ \hline
\end{tabular}

*Attention related, +Stripes Related, \texttildelow  Pose or human pose related
\caption{Comparison with state-of-the-art person Re-ID methods
on the DukeMTMC-ReID dataset.}
\label{tab:duke}
\end{table}

\begin{table}[]
\centering
\resizebox{\linewidth}{!}{
\begin{tabular}{l|l|ll|ll}
\hline
\multirow{2}{*}{Method}                                                & \multirow{2}{*}{Backbone} & \multicolumn{2}{l|}{Labeled} & \multicolumn{2}{l}{Detected} \\ \cline{3-6} 
                                                                       &                           & \textit{mAP}          & \textit{Rank-1}        & \textit{mAP}          & \textit{Rank-1}        \\ \hline
Deep Miner& \texttt{Resnet50}                  & \textbf{84.7 }            & 86.6              & \textbf{81.4}         & 83.5          \\ \hline
$* +$ PLR OSNet \cite{xie2020learning}                                 & \texttt{OSNet}                     & 80.5         & 84.6          & 77.2         & 80.4          \\
$* +$ SCSN \cite{chen2020salience}                                     & \texttt{Resnet50}                  & 84.00        & \textbf{86.80}         & 81.00        & \textbf{84.70}         \\
$+$ Pyramid      \cite{zheng2019pyramidal}                             & \texttt{Resnet101}                 & 76.90        & 78.90         & 74.80        & 78.90         \\
$* +$ MHN \cite{chen2019mixed}                                         & \texttt{Resnet50}                  & 72.40        & 77.20         & 65.40        & 71.70         \\
$+$ MGN \cite{lin2019improving}                                        & \texttt{Resnet50}                  & 67.40        & 68.00         & 66.00        & 68.00         \\
BFE  \cite{dai2019batch}                                               & \texttt{Resnet50}                  & 76.60        & 79.40         & 73.50        & 76.40         \\
$* +$CASN \cite{zheng2019re}                                           & \texttt{Resnet50}                  & 68.00        & 73.70         & 64.40        & 71.50         \\
$+$ PCB+RPP \cite{sun2018beyond}                                       & \texttt{Resnet50}                  & -            & -             & 57.50        & 63.70         \\ \hline
OSNet \cite{zhou2019omni}                                              & \texttt{OSNet}                     & -            & -             & 67.8         & 72.3          \\
Tricks \cite{luo2019bag}                                               & \texttt{SeResnet101}               & 70.40        & 72.00         & 68.00        & 69.60         \\
$*$ Mancs \cite{wang2018mancs}                                         & \texttt{Resnet50}                  & 63.90        & 69.00         & 60.50        & 65.50         \\  \hline
\end{tabular}

}
*Attention related, +Stripes Related, 
\caption{Comparison with state-of-the-art person Re-ID methods
on the the CUHK03 dataset with the 767/700 split.}
\label{tab:cuhk}
\end{table}

\begin{table}[]
\centering
\resizebox{\linewidth}{!}{
\begin{tabular}{l|l|ll}
\hline
Method                                                                 & Backbone    & \textit{mAP}   & \textit{Rank-1} \\ \hline
Deep Miner & \texttt{Resnet50}    & \textbf{67.30 }     & \textbf{85.60 }      \\ \hline
$* +$ SCSN \cite{chen2020salience}                                     & \texttt{Resnet50}    & 58.50 & 83.80  \\
$*$ ABDNet \cite{chen2019abd}                                          & \texttt{Resnet50}    & 60.80 & 82.30  \\
BFE  \cite{dai2019batch}                                               & \texttt{Resnet50}    & 51.50 & 78.80  \\
$*$ IANet \cite{hou2019interaction}                                    & \texttt{Resnet50}    & 46.80 & 75.50  \\
GLAD\cite{wang2018learning}                                                               & \texttt{Resnet50}    & 34.00 & 61.40  \\
PDC\cite{su2017pose}                                                                & \texttt{GoogLeNet}   & 29.70 & 58.00  \\ \hline
\end{tabular}

}
*Attention related, +Stripes Related
\caption{Comparison with state-of-the-art person Re-ID methods
on the MSMT17 dataset.}
\label{tab:msmt17}
\end{table}

Now we compare our proposed Deep Miner model with recent SOTA methods to demonstrate its effectiveness and robustness compared to more advanced methods.

\textbf{Market1501:} Table \ref{tab:market} shows the results on the \textbf{Market1501} dataset.  We divide the methods into two groups: methods that uses local features (top of the table) and methods that uses only global features (bottom). Deep Miner significantly outperforms previous methods in terms of the \textit{mAP} score and has the same \textit{Rank-1} score as the SOTA methods SCSN and Pyramid while being conceptually much simpler than the former.
We also stress that Deep Miner surpasses methods that use stronger backbones while we only consider \texttt{Resnet50}. In fact, authors  in \cite{zheng2019pyramidal} and \cite{alemu2019deep} use \texttt{Resnet101} while authors in \cite{tay2019aanet} and \cite{kalayeh2018human} use a \texttt{Resnet152}. Moreover, Deep Miner -- with a local branch that uses only $4$ stripes -- surpasses more complex methods like Pyramid that uses a pyramidal feature set and a large number of stripes. Overall, the proposed Deep Miner model surpasses all SOTA methods of the global features group demonstrating its ability to mine more diverse and richer features for person re-identification. 

\textbf{DukeMTMC-ReID:} Similar to \textbf{Market1501}, results in Table \ref{tab:duke} show that Deep Miner achieves optimal \textit{mAP} scores compared to all SOTA methods on this dataset.

\textbf{CUHK03:} The results are shown in Table \ref{tab:cuhk}.
Note that this dataset is more challenging than the previous ones in the sense that it contains fewer number of samples and has limited viewpoint variations. 
The proposed Deep Miner model is shown to achieve the optimal \textit{mAP} score compared to SOTA methods. Indeed, Deep Miner exceeds PLR OSNet by $4.2 \%$ in \textit{mAP} on the labeled dataset and exceeds SCSN by $0.7 \%$. On the detected dataset, it surpasses PLR OSNet by $3.2 \%$ and SCSN by $0.4 \%$. 
These experimental results notably express the benefits of the feature mining method, even under the condition of limited training samples.

\textbf{MSMT17:} Lastly, Table \ref{tab:msmt17} shows the results on the \textbf{MSMT17} dataset, which is the more recent and largest dataset. Note that Deep Miner significantly outperforms all SOTA methods in terms of both \textit{mAP} and \textit{Rank-1} scores. Notably, we obtain a \textit{significant gain} of $\textbf{6.5\%}$ \textit{mAP} compared to the best SOTA method on this dataset.

\section{Conclusion}

This paper introduced Deep Miner, a multi-branch network that mines rich and diverse features for person re-identification. This model is composed of three types of branches that are all complementary to each other: the global branch extracts general features of the person; the Input-Erased branches mine richer and more diverse features; the Local branch looks for fine grained details. Specifically, our main insight is to add to a given backbone Input-Erased Branches which take as input partially erased features maps (through a simple erasing strategy) and find new features ignored by the backbone branch. Extensive experiments have demonstrated the effectiveness of the proposed Deep Miner model and its superiority in terms of performance compared to much more complex SOTA methods. It is worth noting that the proposed Deep Miner model makes $\textbf{6.5\%}$ \textit{mAP} improvement on the \textbf{MSMT17} dataset -- the largest and more complex existing person re-identification dataset.

\onecolumn
\begin{multicols}{1}
\vfill\pagebreak
{\small
\bibliographystyle{ieee_fullname}
\bibliography{egbib}

\begin{thebibliography}{10}\itemsep=-1pt

\bibitem{alemu2019deep}
Leulseged~Tesfaye Alemu, Marcello Pelillo, and Mubarak Shah.
\newblock Deep constrained dominant sets for person re-identification.
\newblock In {\em Proceedings of the IEEE International Conference on Computer
  Vision}, pages 9855--9864, 2019.

\bibitem{benassou2020hierarchical}
Sabrina~Narimene Benassou, Wuzhen Shi, Feng Jiang, and Abdallah Benzine.
\newblock Hierarchical complementary learning for weakly supervised object
  localization.
\newblock {\em arXiv preprint arXiv:2011.08014}, 2020.

\bibitem{chen2019energy}
Binghui Chen and Weihong Deng.
\newblock Energy confused adversarial metric learning for zero-shot image
  retrieval and clustering.
\newblock In {\em Proceedings of the AAAI Conference on Artificial
  Intelligence}, volume~33, pages 8134--8141, 2019.

\bibitem{chen2019hybrid}
Binghui Chen and Weihong Deng.
\newblock Hybrid-attention based decoupled metric learning for zero-shot image
  retrieval.
\newblock In {\em Proceedings of the IEEE Conference on Computer Vision and
  Pattern Recognition}, pages 2750--2759, 2019.

\bibitem{chen2019mixed}
Binghui Chen, Weihong Deng, and Jiani Hu.
\newblock Mixed high-order attention network for person re-identification.
\newblock In {\em Proceedings of the IEEE International Conference on Computer
  Vision}, pages 371--381, 2019.

\bibitem{chen2018group}
Dapeng Chen, Dan Xu, Hongsheng Li, Nicu Sebe, and Xiaogang Wang.
\newblock Group consistent similarity learning via deep crf for person
  re-identification.
\newblock In {\em Proceedings of the IEEE Conference on Computer Vision and
  Pattern Recognition}, pages 8649--8658, 2018.

\bibitem{chen2019abd}
Tianlong Chen, Shaojin Ding, Jingyi Xie, Ye Yuan, Wuyang Chen, Yang Yang, Zhou
  Ren, and Zhangyang Wang.
\newblock Abd-net: Attentive but diverse person re-identification.
\newblock In {\em Proceedings of the IEEE International Conference on Computer
  Vision}, pages 8351--8361, 2019.

\bibitem{chen2020salience}
Xuesong Chen, Canmiao Fu, Yong Zhao, Feng Zheng, Jingkuan Song, Rongrong Ji,
  and Yi Yang.
\newblock Salience-guided cascaded suppression network for person
  re-identification.
\newblock In {\em Proceedings of the IEEE/CVF Conference on Computer Vision and
  Pattern Recognition}, pages 3300--3310, 2020.

\bibitem{choe2019attention}
Junsuk Choe and Hyunjung Shim.
\newblock Attention-based dropout layer for weakly supervised object
  localization.
\newblock In {\em Proceedings of the IEEE Conference on Computer Vision and
  Pattern Recognition}, pages 2219--2228, 2019.

\bibitem{dai2019batch}
Zuozhuo Dai, Mingqiang Chen, Xiaodong Gu, Siyu Zhu, and Ping Tan.
\newblock Batch dropblock network for person re-identification and beyond.
\newblock In {\em Proceedings of the IEEE International Conference on Computer
  Vision}, pages 3691--3701, 2019.

\bibitem{felzenszwalb2008discriminatively}
Pedro Felzenszwalb, David McAllester, and Deva Ramanan.
\newblock A discriminatively trained, multiscale, deformable part model.
\newblock In {\em 2008 IEEE conference on computer vision and pattern
  recognition}, pages 1--8. IEEE, 2008.

\bibitem{hermans2017defense}
Alexander Hermans, Lucas Beyer, and Bastian Leibe.
\newblock In defense of the triplet loss for person re-identification.
\newblock {\em arXiv preprint arXiv:1703.07737}, 2017.

\bibitem{hou2019interaction}
Ruibing Hou, Bingpeng Ma, Hong Chang, Xinqian Gu, Shiguang Shan, and Xilin
  Chen.
\newblock Interaction-and-aggregation network for person re-identification.
\newblock In {\em Proceedings of the IEEE Conference on Computer Vision and
  Pattern Recognition}, pages 9317--9326, 2019.

\bibitem{hu2018squeeze}
Jie Hu, Li Shen, and Gang Sun.
\newblock Squeeze-and-excitation networks.
\newblock In {\em Proceedings of the IEEE conference on computer vision and
  pattern recognition}, pages 7132--7141, 2018.

\bibitem{jin2020style}
Xin Jin, Cuiling Lan, Wenjun Zeng, Zhibo Chen, and Li Zhang.
\newblock Style normalization and restitution for generalizable person
  re-identification.
\newblock In {\em Proceedings of the IEEE/CVF Conference on Computer Vision and
  Pattern Recognition}, pages 3143--3152, 2020.

\bibitem{kalayeh2018human}
Mahdi~M Kalayeh, Emrah Basaran, Muhittin G{\"o}kmen, Mustafa~E Kamasak, and
  Mubarak Shah.
\newblock Human semantic parsing for person re-identification.
\newblock In {\em Proceedings of the IEEE Conference on Computer Vision and
  Pattern Recognition}, pages 1062--1071, 2018.

\bibitem{li2014deepreid}
Wei Li, Rui Zhao, Tong Xiao, and Xiaogang Wang.
\newblock Deepreid: Deep filter pairing neural network for person
  re-identification.
\newblock In {\em Proceedings of the IEEE conference on computer vision and
  pattern recognition}, pages 152--159, 2014.

\bibitem{li2018harmonious}
Wei Li, Xiatian Zhu, and Shaogang Gong.
\newblock Harmonious attention network for person re-identification.
\newblock In {\em Proceedings of the IEEE conference on computer vision and
  pattern recognition}, pages 2285--2294, 2018.

\bibitem{lin2019improving}
Yutian Lin, Liang Zheng, Zhedong Zheng, Yu Wu, Zhilan Hu, Chenggang Yan, and Yi
  Yang.
\newblock Improving person re-identification by attribute and identity
  learning.
\newblock {\em Pattern Recognition}, 95:151--161, 2019.

\bibitem{liu2018pose}
Jinxian Liu, Bingbing Ni, Yichao Yan, Peng Zhou, Shuo Cheng, and Jianguo Hu.
\newblock Pose transferrable person re-identification.
\newblock In {\em Proceedings of the IEEE Conference on Computer Vision and
  Pattern Recognition}, pages 4099--4108, 2018.

\bibitem{luo2019bag}
Hao Luo, Youzhi Gu, Xingyu Liao, Shenqi Lai, and Wei Jiang.
\newblock Bag of tricks and a strong baseline for deep person
  re-identification.
\newblock In {\em Proceedings of the IEEE Conference on Computer Vision and
  Pattern Recognition Workshops}, pages 0--0, 2019.

\bibitem{quan2019auto}
Ruijie Quan, Xuanyi Dong, Yu Wu, Linchao Zhu, and Yi Yang.
\newblock Auto-reid: Searching for a part-aware convnet for person
  re-identification.
\newblock In {\em Proceedings of the IEEE International Conference on Computer
  Vision}, pages 3750--3759, 2019.

\bibitem{ren2016faster}
Shaoqing Ren, Kaiming He, Ross Girshick, and Jian Sun.
\newblock Faster r-cnn: Towards real-time object detection with region proposal
  networks, 2016.

\bibitem{ristani2016performance}
Ergys Ristani, Francesco Solera, Roger Zou, Rita Cucchiara, and Carlo Tomasi.
\newblock Performance measures and a data set for multi-target, multi-camera
  tracking.
\newblock In {\em European Conference on Computer Vision}, pages 17--35.
  Springer, 2016.

\bibitem{saquib2018pose}
M Saquib~Sarfraz, Arne Schumann, Andreas Eberle, and Rainer Stiefelhagen.
\newblock A pose-sensitive embedding for person re-identification with expanded
  cross neighborhood re-ranking.
\newblock In {\em Proceedings of the IEEE Conference on Computer Vision and
  Pattern Recognition}, pages 420--429, 2018.

\bibitem{shen2018deep}
Yantao Shen, Hongsheng Li, Tong Xiao, Shuai Yi, Dapeng Chen, and Xiaogang Wang.
\newblock Deep group-shuffling random walk for person re-identification.
\newblock In {\em Proceedings of the IEEE conference on computer vision and
  pattern recognition}, pages 2265--2274, 2018.

\bibitem{si2018dual}
Jianlou Si, Honggang Zhang, Chun-Guang Li, Jason Kuen, Xiangfei Kong, Alex~C
  Kot, and Gang Wang.
\newblock Dual attention matching network for context-aware feature sequence
  based person re-identification.
\newblock In {\em Proceedings of the IEEE Conference on Computer Vision and
  Pattern Recognition}, pages 5363--5372, 2018.

\bibitem{su2017pose}
Chi Su, Jianing Li, Shiliang Zhang, Junliang Xing, Wen Gao, and Qi Tian.
\newblock Pose-driven deep convolutional model for person re-identification.
\newblock In {\em Proceedings of the IEEE international conference on computer
  vision}, pages 3960--3969, 2017.

\bibitem{sun2019perceive}
Yifan Sun, Qin Xu, Yali Li, Chi Zhang, Yikang Li, Shengjin Wang, and Jian Sun.
\newblock Perceive where to focus: Learning visibility-aware part-level
  features for partial person re-identification.
\newblock In {\em Proceedings of the IEEE Conference on Computer Vision and
  Pattern Recognition}, pages 393--402, 2019.

\bibitem{sun2017svdnet}
Yifan Sun, Liang Zheng, Weijian Deng, and Shengjin Wang.
\newblock Svdnet for pedestrian retrieval.
\newblock In {\em Proceedings of the IEEE International Conference on Computer
  Vision}, pages 3800--3808, 2017.

\bibitem{sun2018beyond}
Yifan Sun, Liang Zheng, Yi Yang, Qi Tian, and Shengjin Wang.
\newblock Beyond part models: Person retrieval with refined part pooling (and a
  strong convolutional baseline).
\newblock In {\em Proceedings of the European Conference on Computer Vision
  (ECCV)}, pages 480--496, 2018.

\bibitem{tamaazousti2019universal}
Youssef Tamaazousti, Herv\'e Le~Borgne, C\'eline Hudelot, Mohamed El~Amine
  Seddik, and Mohamed Tamaazousti.
\newblock Learning more universal representations for transfer-learning.
\newblock {\em arXiv:1712.09708}, 2019.

\bibitem{tay2019aanet}
Chiat-Pin Tay, Sharmili Roy, and Kim-Hui Yap.
\newblock Aanet: Attribute attention network for person re-identifications.
\newblock In {\em Proceedings of the IEEE Conference on Computer Vision and
  Pattern Recognition}, pages 7134--7143, 2019.

\bibitem{wang2018mancs}
Cheng Wang, Qian Zhang, Chang Huang, Wenyu Liu, and Xinggang Wang.
\newblock Mancs: A multi-task attentional network with curriculum sampling for
  person re-identification.
\newblock In {\em Proceedings of the European Conference on Computer Vision
  (ECCV)}, pages 365--381, 2018.

\bibitem{wang2020high}
Guan'an Wang, Shuo Yang, Huanyu Liu, Zhicheng Wang, Yang Yang, Shuliang Wang,
  Gang Yu, Erjin Zhou, and Jian Sun.
\newblock High-order information matters: Learning relation and topology for
  occluded person re-identification.
\newblock In {\em Proceedings of the IEEE/CVF Conference on Computer Vision and
  Pattern Recognition}, pages 6449--6458, 2020.

\bibitem{wang2018learning}
Guanshuo Wang, Yufeng Yuan, Xiong Chen, Jiwei Li, and Xi Zhou.
\newblock Learning discriminative features with multiple granularities for
  person re-identification.
\newblock In {\em Proceedings of the 26th ACM international conference on
  Multimedia}, pages 274--282, 2018.

\bibitem{wei2018person}
Longhui Wei, Shiliang Zhang, Wen Gao, and Qi Tian.
\newblock Person transfer gan to bridge domain gap for person
  re-identification.
\newblock In {\em Proceedings of the IEEE Conference on Computer Vision and
  Pattern Recognition}, pages 79--88, 2018.

\bibitem{xie2020learning}
Ben Xie, Xiaofu Wu, Suofei Zhang, Shiliang Zhao, and Ming Li.
\newblock Learning diverse features with part-level resolution for person
  re-identification.
\newblock {\em arXiv preprint arXiv:2001.07442}, 2020.

\bibitem{xu2018attention}
Jing Xu, Rui Zhao, Feng Zhu, Huaming Wang, and Wanli Ouyang.
\newblock Attention-aware compositional network for person re-identification.
\newblock In {\em Proceedings of the IEEE Conference on Computer Vision and
  Pattern Recognition}, pages 2119--2128, 2018.

\bibitem{yi2014deep}
Dong Yi, Zhen Lei, Shengcai Liao, and Stan~Z Li.
\newblock Deep metric learning for person re-identification.
\newblock In {\em 2014 22nd International Conference on Pattern Recognition},
  pages 34--39. IEEE, 2014.

\bibitem{zhang2018adversarial}
Xiaolin Zhang, Yunchao Wei, Jiashi Feng, Yi Yang, and Thomas~S Huang.
\newblock Adversarial complementary learning for weakly supervised object
  localization.
\newblock In {\em Proceedings of the IEEE Conference on Computer Vision and
  Pattern Recognition}, pages 1325--1334, 2018.

\bibitem{zhang2020relation}
Zhizheng Zhang, Cuiling Lan, Wenjun Zeng, Xin Jin, and Zhibo Chen.
\newblock Relation-aware global attention for person re-identification.
\newblock In {\em Proceedings of the IEEE/CVF Conference on Computer Vision and
  Pattern Recognition}, pages 3186--3195, 2020.

\bibitem{zheng2019pyramidal}
Feng Zheng, Cheng Deng, Xing Sun, Xinyang Jiang, Xiaowei Guo, Zongqiao Yu,
  Feiyue Huang, and Rongrong Ji.
\newblock Pyramidal person re-identification via multi-loss dynamic training.
\newblock In {\em Proceedings of the IEEE Conference on Computer Vision and
  Pattern Recognition}, pages 8514--8522, 2019.

\bibitem{zheng2015scalable}
Liang Zheng, Liyue Shen, Lu Tian, Shengjin Wang, Jingdong Wang, and Qi Tian.
\newblock Scalable person re-identification: A benchmark.
\newblock In {\em Proceedings of the IEEE international conference on computer
  vision}, pages 1116--1124, 2015.

\bibitem{zheng2019re}
Meng Zheng, Srikrishna Karanam, Ziyan Wu, and Richard~J Radke.
\newblock Re-identification with consistent attentive siamese networks.
\newblock In {\em Proceedings of the IEEE conference on computer vision and
  pattern recognition}, pages 5735--5744, 2019.

\bibitem{zheng2018pedestrian}
Zhedong Zheng, Liang Zheng, and Yi Yang.
\newblock Pedestrian alignment network for large-scale person
  re-identification.
\newblock {\em IEEE Transactions on Circuits and Systems for Video Technology},
  29(10):3037--3045, 2018.

\bibitem{zhou2019omni}
Kaiyang Zhou, Yongxin Yang, Andrea Cavallaro, and Tao Xiang.
\newblock Omni-scale feature learning for person re-identification.
\newblock In {\em Proceedings of the IEEE International Conference on Computer
  Vision}, pages 3702--3712, 2019.

\end{thebibliography}
}
\end{multicols}
\end{document}